\documentclass[letterpaper]{article} % DO NOT CHANGE THIS
\usepackage{aaai2026}  % DO NOT CHANGE THIS
\usepackage{times}  % DO NOT CHANGE THIS
\usepackage{helvet}  % DO NOT CHANGE THIS
\usepackage{courier}  % DO NOT CHANGE THIS
\usepackage[hyphens]{url}  % DO NOT CHANGE THIS
\usepackage{graphicx} % DO NOT CHANGE THIS
\usepackage{natbib}  % DO NOT CHANGE THIS AND DO NOT ADD ANY OPTIONS TO IT
\usepackage{caption} % DO NOT CHANGE THIS AND DO NOT ADD ANY OPTIONS TO IT
\usepackage{times}
\usepackage{helvet}
\usepackage{courier}
\usepackage{xcolor}

\usepackage{algorithm}
\usepackage{algorithmic}

\usepackage{multirow}
\usepackage{amssymb}
\usepackage{booktabs}
\usepackage[table]{xcolor}
\usepackage{amsmath}

\usepackage{newfloat}
\usepackage{listings}
\DeclareCaptionStyle{ruled}{labelfont=normalfont,labelsep=colon,strut=off} % DO NOT CHANGE THIS
\floatstyle{ruled}
\newfloat{listing}{tb}{lst}{}
\floatname{listing}{Listing}
\title{Tracking the Unstable: Appearance-Guided Motion Modeling for
Robust Multi-Object Tracking in UAV-Captured Videos}
\author{
   Jianbo Ma\textsuperscript{\rm 1,\rm 2}, Hui Luo\textsuperscript{\rm 1}\Corresponding, Qi Chen\textsuperscript{\rm 3}, Yuankai Qi\textsuperscript{\rm 4}, Yumei Sun\textsuperscript{\rm 1}, Amin Beheshti\textsuperscript{\rm 4}, \\ Jianlin Zhang\textsuperscript{\rm 1}\Corresponding, Ming-Hsuan Yang\textsuperscript{\rm 5}
}
\affiliations{
    \textsuperscript{\rm 1}State Key Laboratory of Optical Field Manipulation Science and Technology, Institute of Optics and Electronics, CAS\\
    \textsuperscript{\rm 2}School of Electronic, Electrical and Communication Engineering, UCAS \\
    \textsuperscript{\rm 3}University of Adelaide \\
    \textsuperscript{\rm 4}Macquarie University \\
    \textsuperscript{\rm 5}University of California, Merced\\
    \{majianbo22, luohui19, sunyumei20\}@mails.ucas.ac.cn, \{yuankai.qi, amin.beheshti\}@mq.edu.au,\\ qi.chen04@adelaide.edu.au, jlin@ioe.ac.cn, mhyang@ucmerced.edu
    
}

\begin{document}

\maketitle

\begin{abstract}
Multi-object tracking (MOT) aims to track multiple objects while maintaining consistent identities across frames of a given video. 
In unmanned aerial vehicle (UAV) recorded videos, frequent viewpoint changes and complex UAV-ground relative motion dynamics pose significant challenges, which often lead to unstable affinity measurement and ambiguous association. 
Existing methods typically model motion and appearance cues separately, overlooking their spatio-temporal interplay and resulting in suboptimal tracking performance. 
In this work, we propose AMOT, which jointly exploits appearance and motion cues through two key components: an Appearance-Motion Consistency (AMC) matrix and a Motion-aware Track Continuation (MTC) module.
Specifically, the AMC matrix computes bi-directional spatial consistency under the guidance of appearance features, enabling more reliable and context-aware identity association.
The MTC module complements AMC by reactivating unmatched tracks through appearance-guided predictions that align with Kalman-based predictions, thereby reducing broken trajectories caused by missed detections. 
Extensive experiments on three UAV benchmarks, including VisDrone2019, UAVDT, and VT-MOT-UAV, demonstrate that our AMOT outperforms current state-of-the-art methods and generalizes well in a plug-and-play and training-free manner. 
\end{abstract}

\begin{links}
    \link{Code}{https://github.com/ydhcg-BoBo/AMOT}
\end{links}

\section{Introduction}
Multi-object tracking (MOT) is a fundamental vision task with widespread applications, such as autonomous driving~\cite{zhuang2024robust} and unmanned aerial vehicle (UAV) surveillance~\cite{wu2024temporal}. 
A typical pipeline of MOT is to first detect multiple objects, and then assign each detection to existing tracks through data association, ensuring the continuity of track identities over time. 
Despite progress, robust data association remains challenging, particularly for videos captured by UAV-mounted cameras. 

\begin{figure}[!t]
    \centering
    \includegraphics[width=1.0\linewidth]{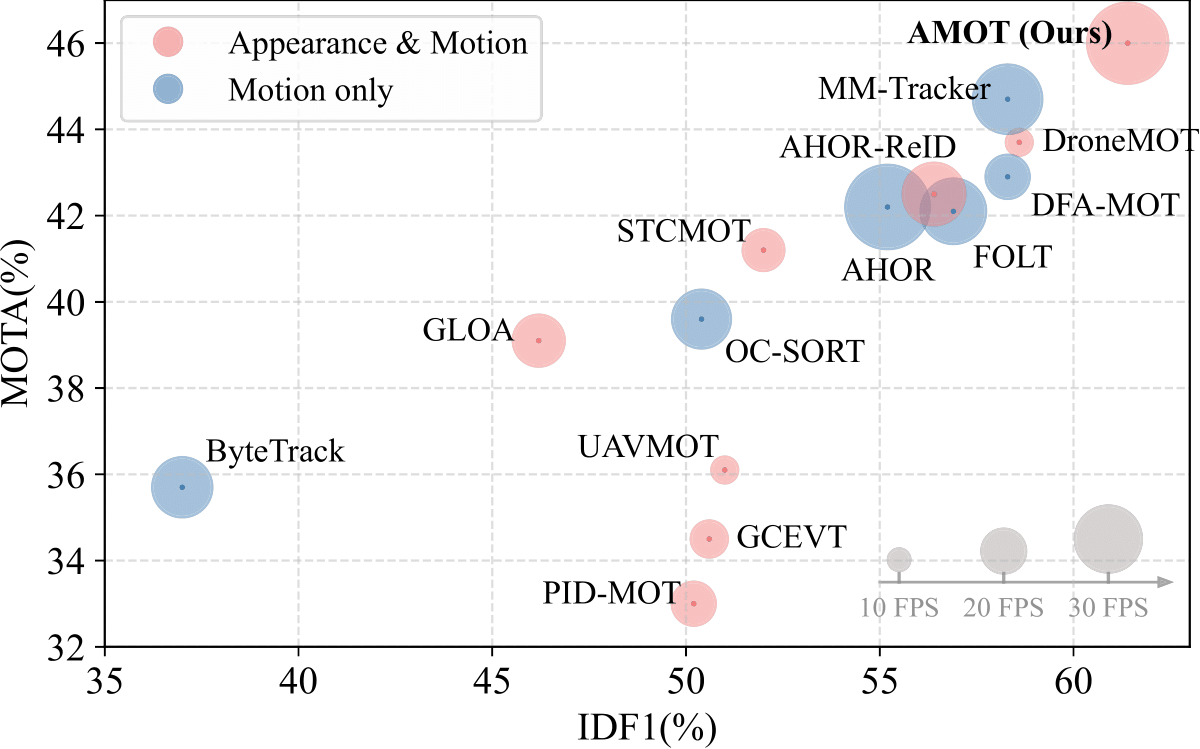}
    \caption{IDF1-MOTA-FPS comparisons of different methods on VisDrone2019. The radius of the circle denotes FPS. Our AMOT achieves the highest IDF1 of 61.4\% and MOTA of 46.0\%, with a real-time inference speed of 36.4 FPS.}
    \label{performance}
\end{figure}

Data association typically relies on a cost matrix that quantifies the affinity between detection–track pairs. 
Frequent viewpoint changes induce significant variations in object appearance and position. 
Compounding this, large and unpredictable object displacements caused by complex UAV-ground relative motion dynamics (e.g., varying velocities and directions) further lead to unstable affinity measurement, ultimately compromising identity assignment. 

To address the above issues, prevailing methods adopt two main strategies for constructing the cost matrix.
(1) Motion-based position prediction: The Kalman Filter~\cite{zhang2022bytetrack,cao2023ocsort} is widely used as a motion model to predict current positions of tracks from their historical states. Then, a motion-based cost matrix is constructed based on the positional proximity between detections and tracks. 
To overcome the limitations of the Kalman Filter's linear motion predictions in UAV views, some recent works introduce complementary techniques, such as camera motion compensation (CMC)~\cite{wang2024dronemot,song2024sftrack} and optical flow~\cite{yao2023folt,yao2024mm-tracker}, to improve the accuracy of predicted track positions. 
(2) Appearance-based instance-level discrimination: 
Instance-level re-identification (ReID) embedding, being insensitive to position changes, offers significant advantages under substantial camera motion or object displacement.
Several works~\cite{li2024matching,wang2024dronemot,wang2024smiletrack,song2024sftrack} have constructed appearance-based cost matrices using ReID embedding to facilitate accurate identity assignment. 
Nevertheless, these two strategies model motion and appearance cues independently to generate separate cost matrices, ignoring the intrinsic relationships between them. 
Specifically, motion prediction errors caused by sudden object displacements can adversely affect the construction of the motion-based cost matrix, while appearance ambiguities similarly impact the appearance-based cost matrix. 
Consequently, when these cost matrices produce conflicting association scores, determining a reliable matching decision becomes difficult.

To address these challenges, we propose an appearance-guided motion modeling strategy that localizes object positions across frames through dense appearance similarity measurement.  
Concretely, we compute a dense response map by measuring the similarity between a query ReID embedding from a reference frame and all spatial locations in the ReID feature map of the adjacent frame.
The response map reflects the probability of the object's spatial position in consecutive frames. 
Building upon this, we introduce an appearance-motion consistency (AMC) matrix that computes forward and backward spatial distances between adjacent frames using dense response maps derived from tracks and detections.
By capturing bi-directional spatial alignment, the AMC matrix reflects strong spatio-temporal correspondence, enabling the construction of a more robust cost matrix. 
In addition, traditional identity assignment depends on detection-to-track matching. 
However, missed detections may cause active tracks to lack corresponding detections, resulting in unmatched tracks. 
To mitigate this issue, we propose a motion-aware track continuation (MTC) module that reactivates unmatched tracks by comparing appearance-guided and Kalman-based object center predictions. 

Together with the AMC and MTC modules, we propose a novel multi-object tracker, termed AMOT, which is built on the joint detection and embedding (JDE) architecture~\cite{zhang2021fairmot}. 
AMOT is specifically designed to enhance the robustness and accuracy of identity assignment under challenging UAV-captured videos. 
Experiments on multiple UAV benchmarks demonstrate that AMOT achieves superior performance in terms of IDF1 and MOTA, as shown in Figure~\ref{performance}. 
The main contributions are summarized as follows:
\begin{itemize}
  \item We propose a novel appearance-motion consistency (AMC) matrix that integrates appearance similarity with bi-directional spatial distances, offering a reliable affinity measurement for robust identity assignment.
  \item We design a motion-aware track continuation (MTC) module that recovers unmatched tracks by aligning appearance-guided and Kalman-based predictions, reducing fragmented trajectories under detection failures. 
  \item AMC and MTC modules are plug-and-play and training-free, allowing easy integration into JDE-based trackers to boost tracking performance. 
  \item Extensive evaluations on VisDrone2019, UAVDT and VT-MOT-UAV benchmarks demonstrate that AMOT consistently outperforms existing trackers. 
  For example, AMOT attains an IDF1 of 61.4\% on VisDrone2019, surpassing MM-Tracker by 2.8\%.
\end{itemize}

\section{Related Works}

\subsection{Motion Modeling} 
The purpose of motion modeling in MOT is to predict the positions of tracks. 
Most MOT methods~\cite{stadler2023improved,li2024sampling,cheng2023dc} adopt the Kalman Filter for motion modeling, owing to its high computational efficiency for real-time applications. 
Predicted positions of tracks are compared with current detections via Mahalanobis distance~\cite{du2023strongsort} or Intersection over Union (IoU)~\cite{lv2024diffmot}, which are employed as motion-based cost metrics for data association. 
Some works~\cite{cao2023ocsort,maggiolino2023deep} enhance the Kalman Filter by adopting an observation-centric strategy to refine track estimation, yielding better performance under non-linear motion patterns. 
Despite their effectiveness, Kalman-based methods exhibit limited capability in scenarios involving substantial camera motion and object displacement. 
Meanwhile, learning-based motion models~\cite{shuai2021siammot, qin2023motiontrack, song2025temporal} have recently gained attention. 
These models leverage data-driven architectures to learn motion patterns. 
However, such methods often incur high computational costs and are unsuitable for real-time tracking. 
In contrast, we introduce a training-free motion modeling strategy that balances robustness and efficiency, specifically tailored for UAV-captured videos. 

\subsection{Appearance Modeling} 
Appearance modeling aims to extract discriminative appearance features to re-identify objects. 
The tracking-by-detection paradigm~\cite{ AHOR,huang2024deconfusetrack} first detects objects using an off-the-shelf detector, and then employs a ReID network to extract identity embedding for each object. 
Although this pipeline delivers impressive performance, it suffers from high computational cost due to separate detection and extraction stages. 
In contrast, the joint detection and embedding (JDE) paradigm~\cite{liu2023uncertainty, meng2023tracking} integrates object detection and ReID feature extraction into a unified framework, enabling simultaneous object localization and embedding extraction. 
Furthermore, several JDE-based trackers incorporate global attention~\cite{GCEVT,GLOA} and temporal cues~\cite{liu2022uavmot, ma2024stcmot} to enhance the discriminability of instance-level ReID embedding. 
The aforementioned methods measure instance-level appearance similarity by computing the cosine distance between ReID embeddings. 
Differently, we reformulate the appearance similarity measurement as a dense response between the instance embedding and the global ReID feature map. This enables the joint modeling of visual similarity and spatial coherence. 

\begin{figure*}[!t]
    \centering
    \includegraphics[width=0.95\linewidth]{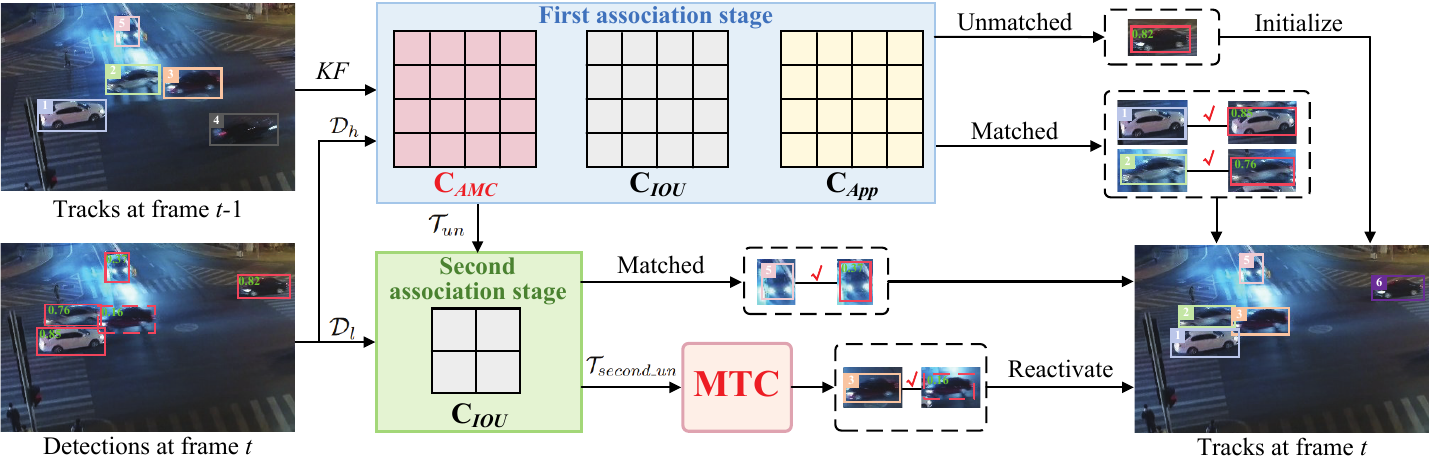}
    \caption{Tracking pipeline of AMOT. 
    Specifically, we introduce the appearance-motion consistency (AMC) matrix $\mathbf{C}_{AMC}$ that integrates it with 
    the appearance similarity matrix $\mathbf{C}_{App}$ and the Intersection-over-Union (IOU) matrix $\mathbf{C}_{IOU}$ to robustly associate high-confidence detections $\mathcal{D}_{h}$ with tracks $\mathcal{T}$ at frame $t-1$. 
    Then, the unmatched tracks $\mathcal{T}_{un}$ are associated with low-confidence detections $\mathcal{D}_{l}$ in the second stage. The remaining unmatched tracks $\mathcal{T}_{second\_un}$ are further potentially reactivated through our proposed motion-aware track continuation (MTC) module. 
    $\textit{KF}$ means the Kalman Filter.}
    \label{fig:pipeline}
\end{figure*} 
\subsection{Data Association} 
Data association typically involves constructing a cost matrix that quantifies the affinity between current detections and existing tracks based on motion and appearance information.
To improve association accuracy, motion and appearance cues are often modeled separately and then integrated into a unified cost matrix~\cite{liang2022one, yang2024hybrid}. 
Once the cost matrix is constructed, data association is formulated as an assignment problem and solved using the Hungarian algorithm~\cite{yi2024ucmctrack, qin2024towards}. 
Nevertheless, existing approaches commonly fail to consider the inherent interplay between motion and appearance cues. 
This independent modeling can cause instability in affinity measurement, arising from prediction errors or appearance ambiguities, which ultimately leads to conflicting data associations. 
To this end, we propose the AMC matrix, which jointly enforces consistency in appearance, motion, and temporal domains, enabling reliable identity assignment.

\section{Methodology}

\subsection{Preliminaries}
Given an input frame $\textbf{I}^{t}$, it is first fed into a feature extractor to obtain a feature map $\textbf{F}^{t}\in \mathbb{R}^{H \times W \times 64}$. 
Then, the detection branch processes $\textbf{F}^{t}$ to generate a heatmap $\textbf{H}^{t} \in \mathbb{R}^{H \times W \times C} $ for object center localization, where C denotes the number of object classes, along with a regression map $\textbf{B}^{t} \in \mathbb{R}^{H \times W \times 2} $ that predicts the height and width of objects. 
The ReID branch produces a ReID feature map $\textbf{E}^{t} \in \mathbb{R}^{  H \times W \times D}$, where $D=128$ is embedding dimension. 

We retain the locations in $\textbf{H}^{t}$ whose confidence scores exceed a threshold $\tau$ as object centers, formulated as:
\begin{align}
    {\mathcal{O}_{det}}=\left \{ \left[x_{i},y_{i}\right]\,\right|\,{\bf H}^{t}_{(x_{i},y_{i})} >\tau\}_{i=1}^{N},
\end{align}
where $\mathcal{O}_{det}$ denotes the set of center coordinates for detections, and $N$ is the total number of detections. 
For each detection, the width and height are obtained from the regression map as $[w_{i},h_{i}]=\textbf{B}^{t}_{(x_{i},y_{i})}$, and the confidence score is given by $c_i=\mathbf{H}^{t}_{(x_{i},y_{i})}$. 
Meanwhile, the ReID embedding $\mathbf{e}_{i}$ for each detection is extracted from $\textbf{E}^{t}$ at the corresponding center coordinate, expressed as $\mathbf{e}_{i}=\mathbf{E}^{t}_{(x_{i},y_{i})}\in\mathbb{R}^{D}$. 

The set of detections is defined as $\mathcal{D}=\{ \mathbf{d}_{i}\}_{i=1}^N$, where each detection is defined as $\mathbf{d}_{i}=[c_{i}, x_{i}, y_{i}, w_{i}, h_{i}, \mathbf{e}_i]$. 
The set of tracks can be represented as $\mathcal{T}=\{\mathbf{t}_{j}\}_{j=1}^M$, where $M$ is the total number of existing tracks. 
The tracking states of a track is defined as $\mathbf{t}_{j}=[id, x_{j}, y_{j}, w_{j}, h_{j}, \mathbf{e}_j]$, where $id$ denotes the track identity, and the set of center coordinates for tracks is defined as:
\begin{align}
    {\mathcal{O}_{trk}}=\{[x_{j},y_{j}]\}_{j=1}^{M}.
\end{align}

In our AMOT, as shown in Figure~\ref{fig:pipeline}, we construct an appearance-motion consistency (AMC) matrix to enable robust detection-to-track association. 
Additionally, we propose a motion-aware track continuation (MTC) module designed to reactivate unmatched tracks without relying on explicit detection-to-track matching. 

\subsection{Appearance-Motion Consistency Matrix}
Most existing cost matrices are constructed using either motion or appearance cues independently. 
However, this separate modeling approach is insufficient under challenging UAV tracking conditions, often resulting in tracking failures. 
To address this issue, we propose the AMC matrix, which jointly models appearance similarity and spatio-temporal correspondence to improve the robustness of association.

Specifically, we compute the track-specific dense response maps $\mathbf{A}_{trk}$ by evaluating the similarity between each track’s ReID embedding and the current ReID feature map $\mathbf{E}^t$, defined as: 
\begin{align}
    \mathbf{A}_{trk}^{(j)}(x, y) = \text{sim}( \mathbf{E}^t(x, y),\,\mathbf{e}_j),
\end{align}
where $\mathbf{e}_j$ denotes the ReID embedding of the $j$-th track, and $\text{sim}(\cdot)$ is the cosine similarity function. 
The response map  $\mathbf{A}_{trk}^{\scriptscriptstyle (j)} \in \mathbb{R}^{H \times W}$ 
highlights regions that are most semantically correlated with the track’s ReID embedding.  
Then, the spatial location corresponding to the maximum response on $\mathbf{A}_{trk}^{ \scriptscriptstyle (j)}$ is defined as the center of the $j$-th track in the current frame. 
Accordingly, the set of predicted center coordinates $\mathcal{Q}_{{trk}}$ for tracks in the current frame is formulated as:
\begin{align}
  \mathcal{Q}_{{trk}}=\{[x_j^*, y_j^*]\,|\,\underset{(x, y) \in \Omega}{\arg\max}\; \mathbf{A}_{trk}^{(j)}(x,y)\}_{j=1}^M.
\end{align} 
Similarly, detection-specific dense response maps $\mathbf{A}_{det}$ are computed 
by evaluating the similarity between each detection’s ReID embedding and the ReID feature map $\mathbf{E}^{t-1}$ from the previous frame, defined as: 
\begin{align}
    \mathbf{A}_{det}^{(i)}(x, y) = \text{sim}(\mathbf{E}^{t-1}(x, y),\,\mathbf{e}_i),
\end{align}
where $\mathbf{e}_i$ denotes the ReID embedding of the  $i$-th detection. 
The set of predicted center coordinates $\mathcal{Q}_{{det}}$ for detections in the previous frame is given by:
\begin{align}
  \mathcal{Q}_{{det}}=\{[x_i^*, y_i^*]\,|\,\underset{(x, y) \in \Omega}{\arg\max}\; \mathbf{A}_{det}^{(i)}(x,y)\}_{i=1}^N.
\end{align}

\begin{figure}[!t]
    \centering
    \includegraphics[width=1.0\linewidth]{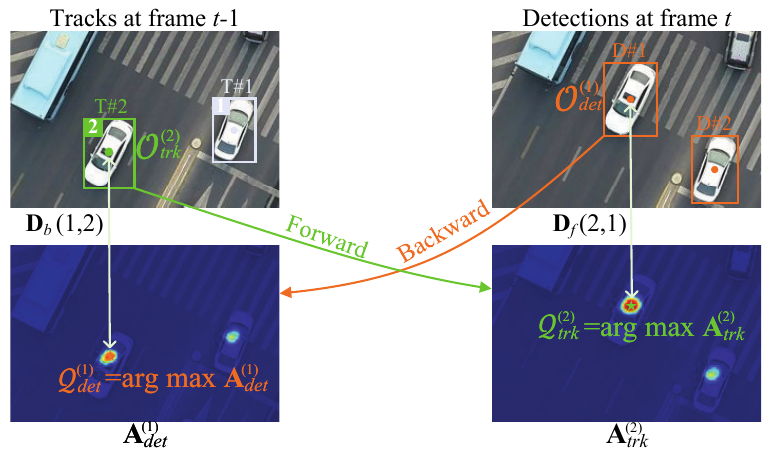}
    \caption{Overview of bi-directional spatial distances in AMC matrix. 
    $\mathbf{A}_{trk}^{\scriptscriptstyle (2)}$ and $\mathbf{A}_{det}^{\scriptscriptstyle (1)}$ are the track-specific dense response map  of T\#2 and the detection-specific dense response map  of D\#1, respectively. $\mathbf{D}_{f}(2,1)$ represents the forward spatial distance from predicted center $\mathcal{Q}_{trk}^{\scriptscriptstyle (2)}$ to observed center $\mathcal{O}_{det}^{\scriptscriptstyle  (1)}$, while $\mathbf{D}_{b}(1,2)$ denotes the backward spatial distance from predicted center $\mathcal{Q}_{det}^{\scriptscriptstyle (1)}$ to observed center $\mathcal{O}_{trk}^{\scriptscriptstyle (2)}$.} 
    \label{fig:model1}
\end{figure}

Subsequently, as shown in Figure~\ref{fig:model1}, we quantify the appearance-guided spatio-temporal correspondence by measuring the forward and backward spatial distances, which can be defined as: 
\begin{align}
    \mathbf{D}_{f}(j,i) = \left\| \mathcal{Q}_{trk}^{(j)}- 
    \mathcal{O}_{det}^{(i)}
    \right\|_2,
\end{align}
\begin{align}
    \mathbf{D}_{b}(i, j) = \left\|\mathcal{Q}_{{det}}^{(i)} - \mathcal{O}_{trk}^{(j)} \right\|_2.
\end{align}
Here, $\mathbf{D}_{f}(j,i)$ denotes the forward spatial distance from the predicted position of the 
$j$-th track $\mathcal{Q}_{trk}^{\scriptscriptstyle(j)}$ to the observed center of the $i$-th detection $\mathcal{O}_{det}^{\scriptscriptstyle(i)}$. 
Conversely, $\mathbf{D}_{b}(i,j)$ represents the backward spatial distance from the predicted position of the $i$-th detection $\mathcal{Q}_{det}^{\scriptscriptstyle(i)}$  to the observed center of the $j$-th track $\mathcal{O}_{trk}^{\scriptscriptstyle(j)}$. 
In both cases, lower values indicate stronger spatial coherence. A detection-track pair is considered a reliable match only when both forward and backward distances are small.

Next, we construct the AMC matrix $\mathbf{C}_{AMC}$ using a Gaussian kernel that integrates the bi-directional spatial distances, defined as:
\begin{align}
    \mathbf{C}_{AMC}(i,j) = 1 - \exp\left(-\frac{\mathbf{D}_{f}^{\top}(i,j) + \mathbf{D}_{b}(i,j)}{2\sigma^2}\right),
\end{align}
where $\sigma$ is a scale factor that controls the spatial sensitivity, and is set to 5. 
$\mathbf{C}_{AMC}$ intrinsically encodes the joint spatial and appearance similarities. 
It is designed to impose a smooth penalty on potentially ambiguous detection-track pairs, enhancing the robustness of affinity measurement.

\subsection{Motion-aware Track Continuation} 
Recovery of short-term lost tracks is critical for maintaining track identities. To this end, we propose the MTC module, which effectively propagates unmatched tracks to mitigate association failures caused by temporary missed detections.

At the beginning of tracking, we initialize a buffer for each track to store its frame indices and corresponding tracking states, denoted as $\text{Buff}=\{\mathbf{t}_j^s\}_{s=t-20}^{t-1}$, and update following a first-in, first-out policy. 
In later frames, if a track fails to associate with any detection, while retaining recent and temporally consecutive tracking states in its buffer, we mark it as a reactivation candidate.
Then, we introduce the MTC module to determine whether these candidates should be reactivated. 
To be specific, we first employ the Kalman Filter to predict the bounding box of all candidates, yielding a set of Kalman-based predictions: $\mathbf{Box}_{kf}=\{[\hat{x}_k,\hat{y}_k,\hat{w}_k,\hat{h}_k]\}_{k=1}^K$, where $K$ is the total number of candidates. 
The corresponding predicted centers are denoted as $\mathbf{c}_{kf}=\{[\hat{x}_k,\hat{y}_k]\}_{k=1}^K$.
Subsequently, we compute the similarity between the candidate’s latest ReID embedding $\mathbf{e}^{t-1}_{k}$ and the current ReID feature map $\mathbf{E}^t$, obtaining a dense response map: $\mathbf{M}^{(k)}=\text{sim}(\mathbf{E}^t,\,\mathbf{e}_{k}^{t-1})\in \mathbb{R}^{H\times W}$. 
The region with the maximum response in $\mathbf{M}$ not only exhibits the highest similarity to the given ReID embedding, but also serves as an appearance-guided prediction of the candidate’s center coordinates in the current frame, denoted as $\mathbf{c}_{reid}=\{[\tilde{x}_k,\tilde{y}_k]\}_{k=1}^K$.

Subsequently, we compute the Euclidean distance between the appearance-guided predicted center $\mathbf{c}_{reid}$ and the Kalman-based predicted center $\mathbf{c}_{kf}$ for each candidate. This distance is formulated as:
\begin{align}
    d_k &= \left\| \mathbf{c}_{reid}^{(k)}-\mathbf{c}_{kf}^{(k)} \right\|_2,
\end{align}
where $d_k$ denotes the spatial offset between the two predicted centers for the $k$-th candidate. 
If $d_k$ is smaller than the predefined threshold $\lambda$, which is set to 3 in our experiments, and there is no significant overlap with any current detection, the candidate is considered present in the current frame and is reactivated to ensure identity consistency. 
Otherwise, it remains unmatched.

\subsection{Tracking Pipeline} 
The tracking pipeline of AMOT is illustrated in Figure~\ref{fig:pipeline}.
The Kalman Filter is used to predict the current positions of tracks. 
We then adopt a two-stage matching strategy similar to BYTE~\cite{zhang2022bytetrack}, where detections are divided into high-confidence sets $\mathcal{D}_h$ and low-confidence sets $\mathcal{D}_l$.

Specifically, in the first stage, we construct three pairwise cost matrices between $\mathcal{D}_h$ and the tracks $\mathcal{T}$, including the appearance similarity matrix $\mathbf{C}_{App}$~\cite{zhang2021fairmot}, the Intersection-over-Union (IOU) matrix $\mathbf{C}_{IOU}$, and the proposed AMC matrix $\mathbf{C}_{AMC}$. 
These matrices are integrated into a unified cost matrix $\mathbf{C}_{uni}$ as follows: 
\begin{align}
    \mathbf{C}_{uni} = 1-(1-\mathbf{C}_{AMC} \cdot \mathbf{C}_{IOU}) \cdot (1- \mathbf{C}_{App}).
\end{align} 
Here, the cost matrix $\mathbf{C}_{uni}$ is used for bipartite matching via the Hungarian algorithm. 
In the second stage, unmatched tracks $\mathcal{T}_{un}$ are associated with $\mathcal{D}_{l}$ solely by the IOU matrix. 
For tracks that remain unmatched after this second association, denoted as $\mathcal{T}_{second\_un}$, we employ the proposed MTC module to determine whether they can be reactivated as matched tracks.

After that, tracks that remain unmatched for more than 30 frames are removed. 
New tracks are initialized from the remaining high-confidence detections that are not associated with any existing tracks, while the tracking states of the matched tracks are updated based on current observations.

\section{Experiments}
\subsection{Settings}
\subsubsection{Datasets and Evaluation Metrics.}
We evaluated our method on various MOT datasets from UAV perspectives, including VisDrone2019~\cite{du2019visdrone}, UAVDT~\cite{du2018uavdt}, and VT-MOT-UAV~\cite{zhu2025vtmot}.

We adopt the widely-used CLEAR metrics~\cite{liu2022uavmot}, including MOTA, IDF1, the number of mostly tracked (MT) objects, and mostly lost (ML) objects, to comprehensively evaluate the tracking performance.  
MOTA emphasizes the detection quality and is computed based on false positives (FP), false negatives (FN), and identity switches (IDs). 
IDF1 measures the tracker’s ability to maintain consistent object identities over time, reflecting the accuracy of identity association across frames. 
Additionally, we report FPS to assess the inference speed of the tracker.

\subsubsection{Implementation Details.} 
We adopt DLA-34~\cite{zhang2021fairmot} as the default backbone network and initialize its parameters with pre-trained weights from the COCO dataset. 
The input image is resized to  $ 608 \times 1088$. 
The corresponding feature map undergoes a 4$\times$ downsampling, resulting in a resolution of $152 \times 272$, where $H=152$ and $W=272$. 
We employ the Adam optimizer with an initial learning rate of $7e^{-5}$ to train our model for 30 epochs, reducing the learning rate by a factor of 10 at 20 epochs. 
For the objective functions, focal loss is employed to supervise the object heatmap, while L1 loss is used to supervise the predicted object width and height. In addition, both cross-entropy loss and triplet loss are applied to guide the learning of the ReID embedding. 
To further enhance the model’s perception of object instances, we introduce an additional mask branch~\cite{tian2020conditional} during training, which is trained using the Dice loss. 

% Our tracker is implemented using Python 3.7 and PyTorch 1.7.1. 
% All experiments are evaluated on an NVIDIA RTX 3090 GPU and Xeon Platinum 8375C 2.90GHz CPU.
% All experiments are conducted on eight NVIDIA RTX 3090 GPUs, and the batch size is set to 32, while tested on a single NVIDIA RTX 3090 GPU.

\subsection{Benchmark Evaluation}
We present the tracking results for multiple UAV datasets. 
$\uparrow$/$\downarrow$ indicate that higher/lower is better, respectively. 
The best scores for each metric are marked in \textbf{bold}, and the second-best scores are marked in \underline{underline}. 

\begin{table}[t]
\small 
\setlength{\tabcolsep}{0.45mm} 
\centering
\begin{tabular}{l|ccccccc}
    \toprule
    Method  &IDF1$\uparrow$ &MOTA$\uparrow$ &MT$\uparrow$ &ML$\downarrow$ &IDs$\downarrow$\\
    \midrule
     ByteTrack~\cite{zhang2022bytetrack}  &37.0 &35.7 &- &-   &2168\\
     UAVMOT~\cite{liu2022uavmot} &51.0 &36.1 &520 &574    &2775\\
     OC-SORT~\cite{cao2023ocsort}  &50.4 &39.6 &- &-  &986\\
     GCEVT~\cite{GCEVT}  &50.6 &34.5 &520 &612 &{841} \\
     GLOA~\cite{GLOA} &46.2 &39.1 &581 &824  &4426\\
     FOLT~\cite{yao2023folt}  &{56.9} &42.1 &- &- &\underline{800}\\
     PID-MOT~\cite{lv2023pid-mot} &50.2 &33.0 &686 &424  &3529\\
     AHOR-ReID~\cite{AHOR} &56.4 &42.5 &- &- &810\\
     STCMOT~\cite{ma2024stcmot}  &{52.0} &{41.2} &{667} &{453}  &3984\\
     DroneMOT~\cite{wang2024dronemot} &\underline{58.6} &{43.7} &\underline{689} &\textbf{397}  &{1112} \\
      DFA-MOT~\cite{DFA-MOT} &58.3 &42.9 &518 &523 &\textbf{792}\\
     MM-Tracker~\cite{yao2024mm-tracker}  &{58.3} &\underline{44.7} &- &- &-\\
     \midrule
     % \rowcolor[HTML]{E6F7FF}
     AMOT (Ours) &\textbf{61.4} &\textbf{46.0} &\textbf{716} &\underline{413}  &1063\\ 
     \bottomrule
\end{tabular}
\caption{Tracking performance comparison with state-of-the-art methods on VisDrone2019 test set.}
\label{VisDrone2019_result}
\end{table}

\begin{table}[t]
\small 
\setlength{\tabcolsep}{0.45mm} 
\centering
\begin{tabular}{l|ccccc}
    \toprule
    Method  &IDF1$\uparrow$ &MOTA$\uparrow$ &MT$\uparrow$ &ML$\downarrow$ &IDs$\downarrow$\\
    \midrule
     ByteTrack~\cite{zhang2022bytetrack} &59.1 &41.6 &- &-  &296\\
     UAVMOT~\cite{liu2022uavmot}  &67.3&46.4 &624 &221   &456 \\
     OC-SORT~\cite{cao2023ocsort}  &64.9 &47.5 &- &-  &{288}\\
     GCEVT~\cite{GCEVT} &68.6 &47.6 &618 &363  &1801 \\
     GLOA~\cite{GLOA}  &68.9 &49.6 &626 &220  &433\\
     FOLT~\cite{yao2023folt} &68.3 &48.5 &- &- &338\\ 
     STCMOT~\cite{ma2024stcmot}  &\underline{69.8} &{49.2} &{664} &{203}  &665\\
     DroneMOT~\cite{wang2024dronemot} &69.6 &{50.1} &638 &\underline{178} &\textbf{129}\\
     DFA-MOT~\cite{DFA-MOT} &69.3 &49.9 &\underline{684} &230 &396 \\
     MM-Tracker~\cite{yao2024mm-tracker}  &{68.9} &\underline{51.4} &- &- &-\\
     % \rowcolor[HTML]{E6F7FF}
     \midrule
     AMOT (Ours) &\textbf{74.7} &\textbf{55.1} &\textbf{794} &\textbf{142} &\underline{272} \\
     \bottomrule
\end{tabular}
\caption{Tracking performance comparison with state-of-the-art methods on UAVDT test set.}
\label{UAVDT_result}
\end{table}

\subsubsection{Results on VisDrone2019.} 
VisDrone2019 serves as a fundamental benchmark for multi-object tracking in dynamic UAV-captured videos, which involve scale variations and long-range object movements. 
As shown in Table~\ref{VisDrone2019_result}, our AMOT exhibits the highest IDF1 of 61.4\% and MOTA of 46.0\%, outperforming current state-of-the-art methods. Concretely, compared with DroneMOT and MM-Tracker, AMOT achieves absolute improvements in both IDF1 and MOTA, demonstrating superior association accuracy and enhanced overall tracking performance.

\subsubsection{Results on UAVDT.} 
The tracking scenarios in UAVDT are derived from bird’s-eye views at different outdoor scenes.
Table~\ref{UAVDT_result} shows that our AMOT achieves the best tracking performance, with an IDF1 of 74.7\% and a MOTA of 55.1\%. 
Specifically, AMOT outperforms STCMOT by +4.9\% in IDF1 and surpasses MM-Tracker by +3.7\% in MOTA, highlighting its superiority in tracking effectiveness.
Moreover, AMOT has the highest MT and the lowest ML, showing its robustness in preserving identity over long-term tracking.

\subsubsection{Results on VT-MOT-UAV.}
VT-MOT-UAV is dedicated to tracking multiple categories, including pedestrians and vehicles, and presents challenges such as varying illumination conditions and cluttered backgrounds. 
The results are summarized in Table~\ref{VTMOT—UAV_result}. Our AMOT exhibits superior performance with an IDF1 of 52.7\% and a MOTA of 31.8\%, surpassing the previous advanced approaches.

\begin{table}[t]
\small 
\setlength{\tabcolsep}{0.6mm} 
\centering
\begin{tabular}{l|ccccc}
    \toprule
    Method  &IDF1$\uparrow$ &MOTA$\uparrow$ &MT$\uparrow$ &ML$\downarrow$  &IDs$\downarrow$\\
    \midrule
     SORT~\cite{bewley2016simple} &48.1 &28.6 &103 &340  &520 \\
     FairMOT~\cite{zhang2021fairmot} &39.9 &17.9 &94 &322&901 \\
     ByteTrack~\cite{zhang2022bytetrack} &\underline{50.2} &\underline{28.5} &129 & 324 &\underline{415}\\
     UAVMOT~\cite{liu2022uavmot} &{47.9} &22.1 &\textbf{175} &\underline{248}  &1421\\
     OC-SORT~\cite{cao2023ocsort} &{48.2} &\underline{28.5} &102 &342  &509 \\
     STCMOT~\cite{ma2024stcmot}  &47.4 &{27.9} &153 &277 &{518}\\
     \midrule
     % \rowcolor[HTML]{E6F7FF}
     AMOT (Ours) &\textbf{52.7} &\textbf{31.8} &\underline{168} &\textbf{247} &\textbf{412}\\
     \bottomrule
\end{tabular}
\caption{Tracking performance comparison with state-of-the-art methods on VT-MOT-UAV test set.}
\label{VTMOT—UAV_result}
\end{table}

\begin{table}[!t]
\small 
\setlength{\tabcolsep}{0.6mm} 
\centering
\begin{tabular}{cc|ccc|ccc|c}
    \toprule
    \multirow{2}{*}{AMC} & \multirow{2}{*}{MTC}  &\multicolumn{3}{c|}{VisDrone2019} &\multicolumn{3}{c|}{UAVDT}\\
    \cline{3-9}
     &  &IDF1$\uparrow$    &MOTA$\uparrow$ &IDs$\downarrow$ &IDF1$\uparrow$  &MOTA$\uparrow$ &IDs$\downarrow$  &FPS $\uparrow$\\
    \midrule
    & 
    &54.4 &43.3 &3847 &72.4 &54.6 &886  &37.1\\
    \checkmark & 
    &60.5 &44.1 &1078 &74.4 &54.9 &307  &36.3\\
    &\checkmark 
    &57.3 &44.2 &1961 &73.4 &54.6 &440  &\textbf{37.7}\\
    \checkmark &\checkmark 
    &\textbf{61.4} &\textbf{46.0} &\textbf{1063} &\textbf{74.7} &\textbf{55.1} &\textbf{272}  &36.4\\
    \bottomrule
\end{tabular}
\caption{Ablation of various components.}
\label{ablation_result}
\end{table}

\subsection{Ablation Studies}

\subsubsection{Baseline model.}
The baseline model uses the network architecture of FairMOT~\cite{zhang2021fairmot} combined with a two-stage association strategy~\cite{zhang2022bytetrack}.

\begin{table}[t]
\small 
\setlength{\tabcolsep}{1.3mm} 
\centering
\begin{tabular}{cc|ccc|ccc}
    \toprule
   \multicolumn{2}{c|}{AMC}  &\multicolumn{3}{c|}{VisDrone2019} &\multicolumn{3}{c}{UAVDT}\\
    \cline{1-8}
    For. &Back. &IDF1$\uparrow$  &MOTA$\uparrow$ &IDs$\downarrow$ &IDF1$\uparrow$  &MOTA$\uparrow$ &IDs$\downarrow$  \\
    \midrule
    & 
    &54.4 &43.3 &3847 &72.4 &54.6 &886\\
    \checkmark &  
    &60.4 &43.4 &1319 &73.3 &54.9 &389\\
    &\checkmark  
    &60.0 &43.9 &1132 &73.0 &54.9 &407 \\
    \checkmark &\checkmark 
    &\textbf{60.5} &\textbf{44.1} &\textbf{1078} &\textbf{74.4} &\textbf{54.9} &\textbf{307}\\
    \bottomrule
\end{tabular}
\caption{Impact of the forward and backward spatial distance in AMC matrix, where “For.” and “Back.” denote forward and backward spatial distances, respectively.}
\label{AMC_ablation_result}
\end{table}

\begin{figure}[!t]
    \centering
    \includegraphics[width=1.0\linewidth]{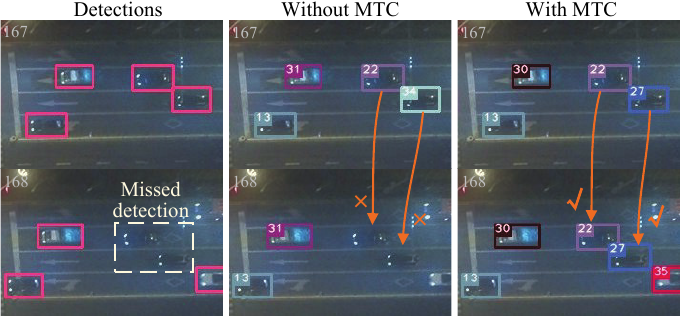}
    \caption{Visualization of tracking results with the MTC module. Despite missed detections, MTC effectively propagates tracks and maintains their correct identities.}
    \label{vis_mtc}
\end{figure}

\begin{figure}[!t]
    \centering
    \includegraphics[width=1.0\linewidth]{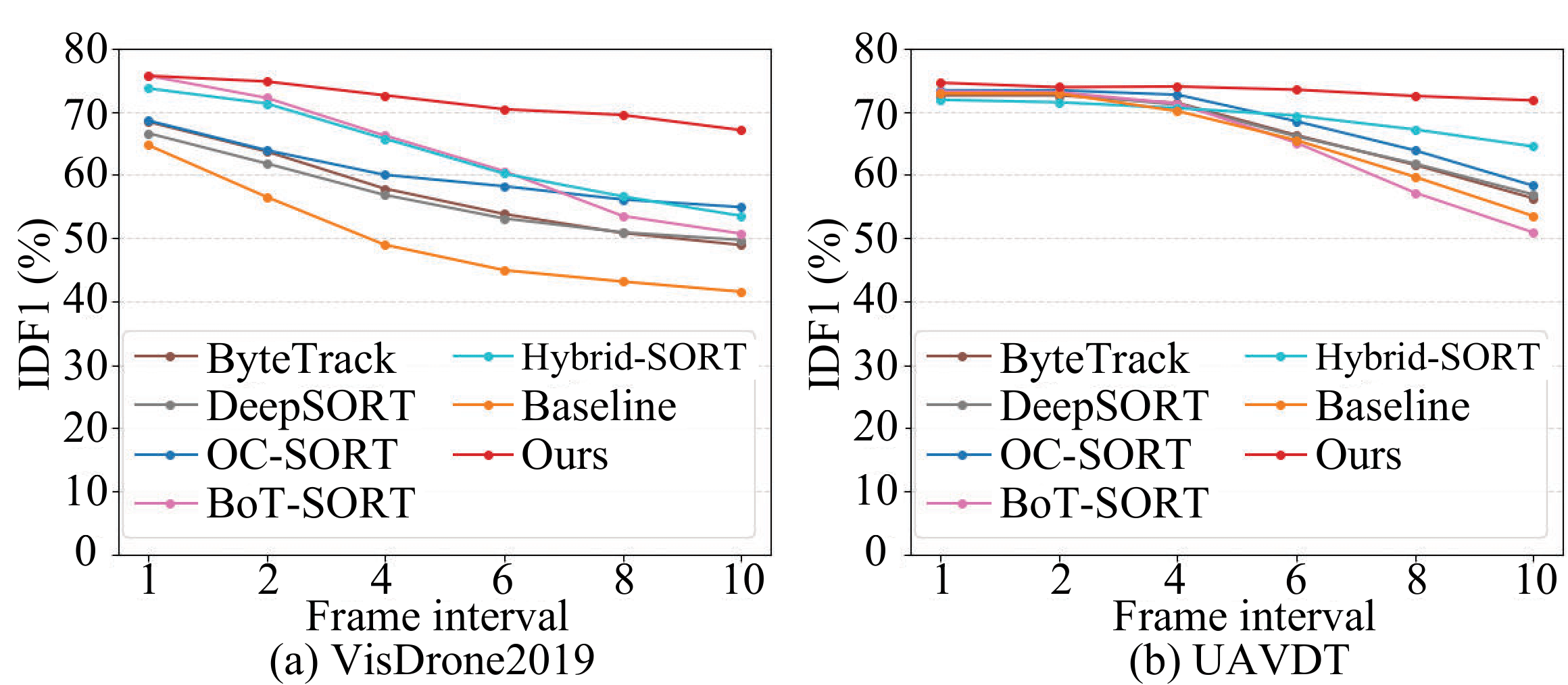}
    \caption{Association performance comparison under different frame intervals for car category. Object displacement increases with frame interval.}
    \label{Diff_trackers}
\end{figure}

\subsubsection{Component Ablation.}
We conduct a comprehensive evaluation of the proposed components, as presented in Table~\ref{ablation_result}. 
Integrating either AMC or MTC component into the baseline model individually leads to notable improvements in tracking performance. 
Compared with the baseline model, the joint application of AMC and MTC improves IDF1 and MOTA by +7.0\% and +2.7\%, respectively, while reducing identity switches by 2784 on VisDrone2019. On UAVDT, it achieves gains of +2.3\% in IDF1 and +0.5\% in MOTA, with 614 fewer identity switches. 
Furthermore, the AMC and MTC introduce only a minimal computational overhead and have a negligible impact on the overall inference speed.
Specifically, the baseline model achieves an inference speed of 37.1 FPS, while our AMOT achieves a comparable speed of 36.4 FPS.
These results highlight the effectiveness and computational efficiency of the proposed components.

\subsubsection{Impact of Bi-directional Spatial Distances.}
As shown in Table~\ref{AMC_ablation_result}, we assess the effect of forward and backward spatial distances in the AMC matrix. 
Both forward and backward spatial distances positively contribute to overall tracking performance, while their joint integration yields further gains. 
This indicates that modeling bi-directional spatial consistency strengthens the spatio-temporal interplay between motion and appearance cues, thereby improving the discriminative power of the affinity measurement.

\begin{figure*}[!t]
    \centering
    \includegraphics[width=1.\linewidth]{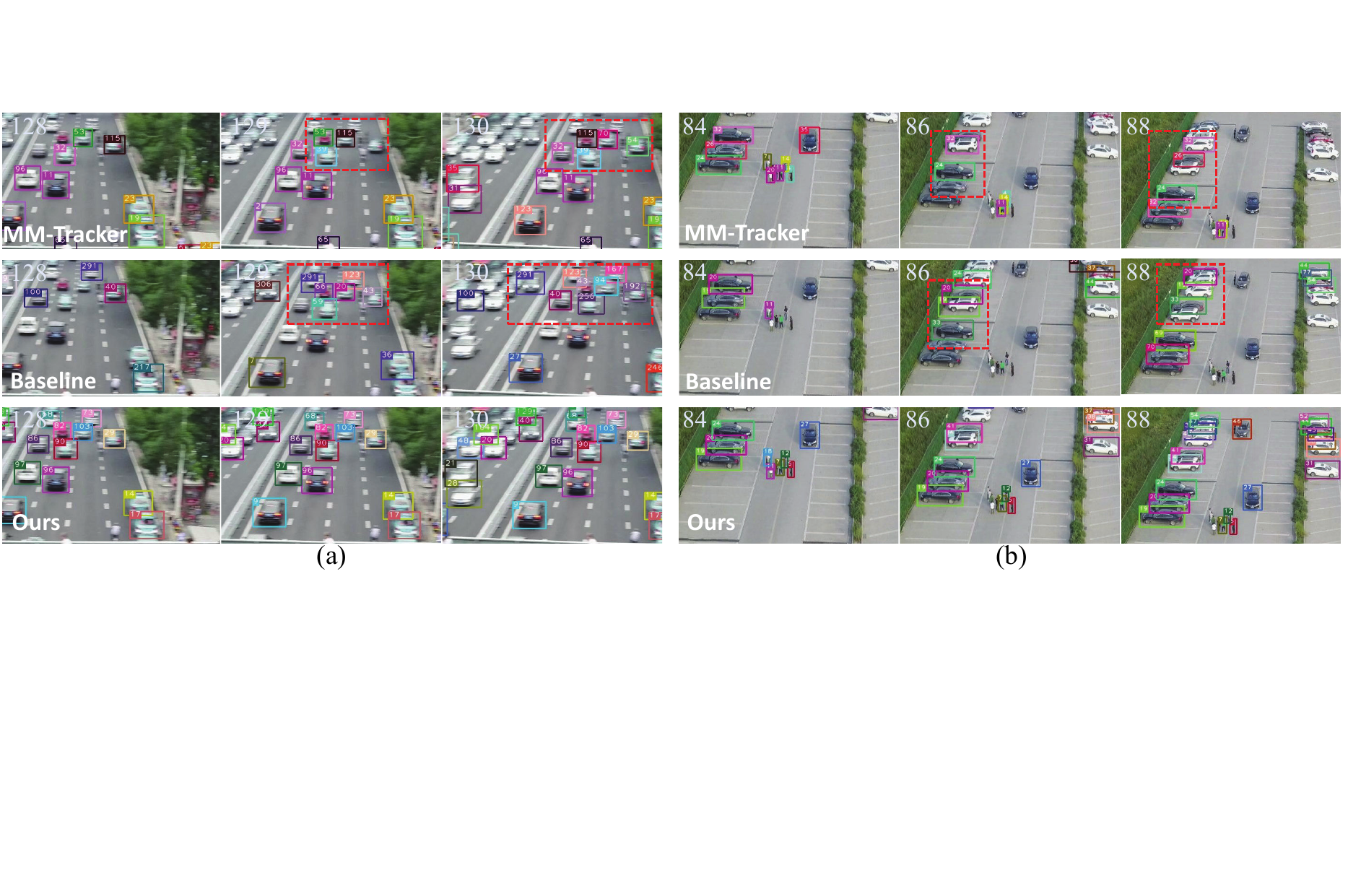}
    \caption{Visualization of tracking results under challenging UAV-captured videos: (a) lateral viewpoint changes from left to right and (b) viewpoint changes from close-up to distant view.}
    \label{fig:compare}
\end{figure*}

\subsubsection{Effect of MTC.}
Figure~\ref{vis_mtc} presents a qualitative comparison of tracking results with and without the proposed MTC module. 
At frame 168, the baseline model fails to associate the object due to missed detections, leading to a track discontinuity. 
In contrast, the baseline model with MTC module successfully propagates the tracks from frame 167 to frame 168 with consistent identity assignment, highlighting the effectiveness of MTC in preserving track continuity under unreliable detection conditions.

\subsubsection{Robustness to Large Displacements.}
In Figure~\ref{Diff_trackers}, we compare the association performance of our AMOT against several advanced data association methods using the same detections under varying frame intervals.
As the frame interval increases,  inter-frame object displacements become larger, resulting in a significant decline in association accuracy for most methods. 
In contrast, our AMOT demonstrates strong robustness under these challenging conditions. 
Although recent competitive methods such as BoT-SORT~\cite{aharon2022bot} and Hybrid-SORT~\cite{yang2024hybrid} exploit both motion and appearance cues for association, they model these cues independently, which limits their ability to handle large object displacements. 
Instead, our method jointly and adaptively models motion and appearance cues, enabling more consistent and accurate data association in highly dynamic tracking scenarios.

\begin{table}[t]
\centering
\small 
\setlength{\tabcolsep}{0.2mm} 
\begin{tabular}{l|cc|ccc|ccc}
    \toprule
    \multirow{2}{*}{Method} &\multirow{2}{*}{AMC} &\multirow{2}{*}{MTC}   &\multicolumn{3}{c|}{VisDrone2019} &\multicolumn{3}{c}{UAVDT}\\
    \cline{4-9}
    & &  &IDF1$\uparrow$  &MOTA$\uparrow$ &IDs$\downarrow$ &IDF1$\uparrow$  &MOTA$\uparrow$ &IDs$\downarrow$\\
    \midrule
    \multirow{3}{*}{\shortstack[l]{UAVMOT }} & & 
    &50.6 &40.0 &3843 &69.0 &47.6 &2079\\
    &\checkmark & 
    &53.7 &40.2 &1589 &69.5 &47.6 &846\\
    &\checkmark &\checkmark 
    &55.0 &42.4 &1600 &69.7 &47.8 &729\\
    \cline{2-9}
    &\multicolumn{2}{c|}{Gain}
    &\textbf{+4.4} &\textbf{+2.4} &\textbf{-2243} &\textbf{+0.7} &\textbf{+0.2} &\textbf{-1350}\\
    \midrule
    \multirow{3}{*}{\shortstack[l]{STCMOT }} & &
    &52.7 &41.0 &4040 &69.1 &50.4 &688\\
    &\checkmark & 
    &58.9  &41.2 &1004 &69.7 &50.6 &352\\
    &\checkmark &\checkmark
    &59.7 &43.1 &1043 &70.0 &50.8 &308\\
    \cline{2-9}
    &\multicolumn{2}{c|}{Gain}
    &\textbf{+7.0} &\textbf{+2.1} &\textbf{-2997} &\textbf{+0.9} &\textbf{+0.4} &\textbf{-380}\\
    \bottomrule
\end{tabular}
\caption{Results of applying AMC and MTC components to various JDE-based methods.}
\label{other_tracker_result}
\end{table}

\subsubsection{Generality on Other Trackers.} 
We apply our design to the representative JDE-based trackers, including UAVMOT~\cite{liu2022uavmot} and STCMOT~\cite{ma2024stcmot}. 
As presented in Table~\ref{other_tracker_result}, all of these trackers benefit from the integration of AMC and MTC. 
For example, STCMOT achieves a notable gain of +7.0\% IDF1 and +2.1\% MOTA on VisDrone2019. 
Similar performance improvements are observed for  UAVMOT, with IDF1 and MOTA increasing by +4.4\% and +2.4\%, respectively. 
Such improvements across diverse trackers and scenarios further demonstrate the generality and effectiveness of our approach. 
% More importantly, the AMC and MTC modules are training-free and plug-and-play, enabling seamless integration into existing MOT frameworks to improve tracking performance in complex UAV-based scenarios.

\subsection{Visualization} 
Frequent viewpoint changes in UAV-captured videos often induce substantial variations in object appearance and position, posing significant challenges to multi-object trackers. 
As illustrated in Figure~\ref{fig:compare}, we evaluate the tracking performance of our method under these adverse conditions. 
Both the baseline model and MM-tracker~\cite{yao2024mm-tracker} suffer from frequent identity switches and tracking failures, reflecting limited capability in coping with dynamic viewpoint changes. 
In contrast, the proposed AMOT demonstrates robust performance and maintains track identity consistency under such challenging conditions.

\section{Conclusion} 
In this work, we explore the joint modeling of appearance and motion cues, and introduce two plug-and-play components for data association, namely the AMC matrix and the MTC module.
AMC captures appearance, motion, and temporal coherence to ensure accurate identity assignment, while MTC mitigates broken trajectories by reactivating unmatched tracks after missed detections. 
Building upon these components, we develop AMOT, a simple yet effective joint detection and embedding framework tailored for real-time UAV tracking. 
Experiments demonstrate that AMOT exhibits strong robustness in
dynamic scenarios captured by UAV-mounted cameras. 
It also achieves outstanding performance on several UAV benchmarks, including VisDrone2019, UAVDT, and VT-MOT-UAV. 

\section{Acknowledgments}
We gratefully acknowledge the support of Prof. Haorui Zuo. 
The work was partially supported by Frontier Research Fund of the Institute of Optics and Electronics, Chinese Academy of Sciences (C24K003).

\appendix 
\bibliography{aaai2026}

\begin{thebibliography}{40}
\providecommand{\natexlab}[1]{#1}

\bibitem[{Aharon et~al.(2022)Aharon, Orfaig, Bobrovsky et~al.}]{aharon2022bot}
Aharon, N.; Orfaig, R.; Bobrovsky, B.-Z.; et~al. 2022.
\newblock BoT-SORT: Robust associations multi-pedestrian tracking.
\newblock \emph{arXiv preprint arXiv:2206.14651}.

\bibitem[{Bewley et~al.(2016)Bewley, Ge, Ott, Ramos, and Upcroft}]{bewley2016simple}
Bewley, A.; Ge, Z.; Ott, L.; Ramos, F.; and Upcroft, B. 2016.
\newblock Simple online and realtime tracking.
\newblock In \emph{IEEE international conference on image processing}, 3464--3468.

\bibitem[{Cao et~al.(2023)Cao, Pang, Weng, Khirodkar, and Kitani}]{cao2023ocsort}
Cao, J.; Pang, J.; Weng, X.; Khirodkar, R.; and Kitani, K. 2023.
\newblock Observation-centric sort: Rethinking sort for robust multi-object tracking.
\newblock In \emph{IEEE/CVF conference on computer vision and pattern recognition}, 9686--9696.

\bibitem[{Cheng et~al.(2023)Cheng, Yao, Xiao, and other}]{cheng2023dc}
Cheng, S.; Yao, M.; Xiao, X.; and other. 2023.
\newblock Dc-mot: Motion deblurring and compensation for multi-object tracking in uav videos.
\newblock In \emph{IEEE International Conference on Robotics and Automation}, 789--795.

\bibitem[{Du et~al.(2018)Du, Qi, Yu, Yang, Duan, Li, Zhang, Huang, and Tian}]{du2018uavdt}
Du, D.; Qi, Y.; Yu, H.; Yang, Y.; Duan, K.; Li, G.; Zhang, W.; Huang, Q.; and Tian, Q. 2018.
\newblock The unmanned aerial vehicle benchmark: Object detection and tracking.
\newblock In \emph{European conference on computer vision}, 370--386.

\bibitem[{Du et~al.(2019)Du, Zhu, Wen, Bian, Lin, Hu, Peng, Zheng, Wang, Zhang et~al.}]{du2019visdrone}
Du, D.; Zhu, P.; Wen, L.; Bian, X.; Lin, H.; Hu, Q.; Peng, T.; Zheng, J.; Wang, X.; Zhang, Y.; et~al. 2019.
\newblock VisDrone-DET2019: The vision meets drone object detection in image challenge results.
\newblock In \emph{IEEE/CVF international conference on computer vision workshops}, 213–226.

\bibitem[{Du et~al.(2023)Du, Zhao, Song, Zhao, Su, Gong, and Meng}]{du2023strongsort}
Du, Y.; Zhao, Z.; Song, Y.; Zhao, Y.; Su, F.; Gong, T.; and Meng, H. 2023.
\newblock Strongsort: Make deepsort great again.
\newblock \emph{IEEE Transactions on Multimedia}, 25: 8725--8737.

\bibitem[{Huang et~al.(2024)Huang, Han, He, Zheng, and Wei}]{huang2024deconfusetrack}
Huang, C.; Han, S.; He, M.; Zheng, W.; and Wei, Y. 2024.
\newblock Deconfusetrack: Dealing with confusion for multi-object tracking.
\newblock In \emph{Proceedings of the IEEE/CVF Conference on Computer Vision and Pattern Recognition}, 19290--19299.

\bibitem[{Jin et~al.(2024)Jin, Nie, Yan, Chen, Zhu, and Qi}]{AHOR}
Jin, H.; Nie, X.; Yan, Y.; Chen, X.; Zhu, Z.; and Qi, D. 2024.
\newblock AHOR: Online Multi-Object Tracking With Authenticity Hierarchizing and Occlusion Recovery.
\newblock \emph{IEEE Transactions on Circuits and Systems for Video Technology}, 34(9): 8253--8265.

\bibitem[{Li et~al.(2024{\natexlab{a}})Li, Ke, Danelljan, Piccinelli, Segu, Van~Gool, and Yu}]{li2024matching}
Li, S.; Ke, L.; Danelljan, M.; Piccinelli, L.; Segu, M.; Van~Gool, L.; and Yu, F. 2024{\natexlab{a}}.
\newblock Matching anything by segmenting anything.
\newblock In \emph{Proceedings of the IEEE/CVF Conference on Computer Vision and Pattern Recognition}, 18963--18973.

\bibitem[{Li et~al.(2024{\natexlab{b}})Li, Zhang, Wu, Song, and Chen}]{li2024sampling}
Li, Z.; Zhang, D.; Wu, S.; Song, M.; and Chen, G. 2024{\natexlab{b}}.
\newblock Sampling-resilient multi-object tracking.
\newblock In \emph{Proceedings of the AAAI Conference on Artificial Intelligence}, 3297--3305.

\bibitem[{Liang et~al.(2022)Liang, Zhang, Zhou, Li, and Hu}]{liang2022one}
Liang, C.; Zhang, Z.; Zhou, X.; Li, B.; and Hu, W. 2022.
\newblock One more check: making “fake background” be tracked again.
\newblock In \emph{Proceedings of the AAAI Conference on Artificial Intelligence}, 1546--1554.

\bibitem[{Liu et~al.(2023)Liu, Jin, Fu, Chen, Jiang, and Ye}]{liu2023uncertainty}
Liu, K.; Jin, S.; Fu, Z.; Chen, Z.; Jiang, R.; and Ye, J. 2023.
\newblock Uncertainty-aware unsupervised multi-object tracking.
\newblock In \emph{Proceedings of the IEEE/CVF international conference on computer vision}, 9996--10005.

\bibitem[{Liu et~al.(2022)Liu, Li, Lu, and He}]{liu2022uavmot}
Liu, S.; Li, X.; Lu, H.; and He, Y. 2022.
\newblock Multi-Object Tracking Meets Moving UAV.
\newblock In \emph{IEEE/CVF Conference on Computer Vision and Pattern Recognition}, 8866--8875.

\bibitem[{Lv et~al.(2024{\natexlab{a}})Lv, Huang, Zhang, Lin, Han, and Zeng}]{lv2024diffmot}
Lv, W.; Huang, Y.; Zhang, N.; Lin, R.-S.; Han, M.; and Zeng, D. 2024{\natexlab{a}}.
\newblock Diffmot: A real-time diffusion-based multiple object tracker with non-linear prediction.
\newblock In \emph{Proceedings of the IEEE/CVF conference on computer vision and pattern recognition}, 19321--19330.

\bibitem[{Lv et~al.(2024{\natexlab{b}})Lv, Zhang, Zhang, and Zeng}]{lv2023pid-mot}
Lv, W.; Zhang, N.; Zhang, J.; and Zeng, D. 2024{\natexlab{b}}.
\newblock One-shot multiple object tracking with robust id preservation.
\newblock \emph{IEEE Transactions on Circuits and Systems for Video Technology}, 34(6): 4473--4488.

\bibitem[{Ma et~al.(2024)Ma, Tang, Wu, Zhao, Zhang, and Xu}]{ma2024stcmot}
Ma, J.; Tang, C.; Wu, F.; Zhao, C.; Zhang, J.; and Xu, Z. 2024.
\newblock STCMOT: Spatio-Temporal Cohesion Learning for UAV-Based Multiple Object Tracking.
\newblock In \emph{IEEE International Conference on Multimedia and Expo}, 1--6.

\bibitem[{Maggiolino et~al.(2023)Maggiolino, Ahmad, Cao, and Kitani}]{maggiolino2023deep}
Maggiolino, G.; Ahmad, A.; Cao, J.; and Kitani, K. 2023.
\newblock Deep oc-sort: Multi-pedestrian tracking by adaptive re-identification.
\newblock In \emph{IEEE International conference on image processing}, 3025--3029.

\bibitem[{Meng et~al.(2023)Meng, Shao, Guo, and Gao}]{meng2023tracking}
Meng, S.; Shao, D.; Guo, J.; and Gao, S. 2023.
\newblock Tracking without label: Unsupervised multiple object tracking via contrastive similarity learning.
\newblock In \emph{Proceedings of the IEEE/CVF international conference on computer vision}, 16264--16273.

\bibitem[{Qin et~al.(2024)Qin, Wang, Zhou, Fu, Hua, and Tang}]{qin2024towards}
Qin, Z.; Wang, L.; Zhou, S.; Fu, P.; Hua, G.; and Tang, W. 2024.
\newblock Towards generalizable multi-object tracking.
\newblock In \emph{Proceedings of the IEEE/CVF Conference on Computer Vision and Pattern Recognition}, 18995--19004.

\bibitem[{Qin et~al.(2023)Qin, Zhou, Wang, Duan, Hua, and Tang}]{qin2023motiontrack}
Qin, Z.; Zhou, S.; Wang, L.; Duan, J.; Hua, G.; and Tang, W. 2023.
\newblock Motiontrack: Learning robust short-term and long-term motions for multi-object tracking.
\newblock In \emph{Proceedings of the IEEE/CVF conference on computer vision and pattern recognition}, 17939--17948.

\bibitem[{Shi et~al.(2023)Shi, Zhang, Pan, Zhang, and Su}]{GLOA}
Shi, L.; Zhang, Q.; Pan, B.; Zhang, J.; and Su, Y. 2023.
\newblock Global-Local and Occlusion Awareness Network for Object Tracking in UAVs.
\newblock \emph{IEEE Journal of Selected Topics in Applied Earth Observations and Remote Sensing}, 16: 8834--8844.

\bibitem[{Shuai et~al.(2021)Shuai, Berneshawi, Li, Modolo, and Tighe}]{shuai2021siammot}
Shuai, B.; Berneshawi, A.; Li, X.; Modolo, D.; and Tighe, J. 2021.
\newblock Siammot: Siamese multi-object tracking.
\newblock In \emph{Proceedings of the IEEE/CVF conference on computer vision and pattern recognition}, 12372--12382.

\bibitem[{Song and Lee(2024)}]{song2024sftrack}
Song, I.; and Lee, J. 2024.
\newblock SFTrack: A Robust Scale and Motion Adaptive Algorithm for Tracking Small and Fast Moving Objects.
\newblock In \emph{IEEE/RSJ International Conference on Intelligent Robots and Systems}, 10870--10877.

\bibitem[{Song et~al.(2025)Song, Luo, Ma, Tang, Chen, Yu, and Yang}]{song2025temporal}
Song, Z.; Luo, R.; Ma, L.; Tang, Y.; Chen, Y.-P.~P.; Yu, J.; and Yang, W. 2025.
\newblock Temporal Coherent Object Flow for Multi-Object Tracking.
\newblock In \emph{Proceedings of the AAAI Conference on Artificial Intelligence}, 6978--6986.

\bibitem[{Stadler and Beyerer(2023)}]{stadler2023improved}
Stadler, D.; and Beyerer, J. 2023.
\newblock An improved association pipeline for multi-person tracking.
\newblock In \emph{Proceedings of the IEEE/CVF conference on computer vision and pattern recognition}, 3170--3179.

\bibitem[{Tian, Shen, and Chen(2020)}]{tian2020conditional}
Tian, Z.; Shen, C.; and Chen, H. 2020.
\newblock Conditional convolutions for instance segmentation.
\newblock In \emph{European conference on computer vision}, 282--298.

\bibitem[{Wang et~al.(2024{\natexlab{a}})Wang, Wang, Li et~al.}]{wang2024dronemot}
Wang, P.; Wang, Y.; Li, D.; et~al. 2024{\natexlab{a}}.
\newblock DroneMOT: Drone-based Multi-Object Tracking Considering Detection Difficulties and Simultaneous Moving of Drones and Objects.
\newblock In \emph{IEEE International Conference on Robotics and Automation}, 7397--7404.

\bibitem[{Wang et~al.(2024{\natexlab{b}})Wang, Hsieh, Chen, Chang, So, and Li}]{wang2024smiletrack}
Wang, Y.-H.; Hsieh, J.-W.; Chen, P.-Y.; Chang, M.-C.; So, H.-H.; and Li, X. 2024{\natexlab{b}}.
\newblock Smiletrack: Similarity learning for occlusion-aware multiple object tracking.
\newblock In \emph{Proceedings of the AAAI conference on artificial intelligence}, 5740--5748.

\bibitem[{Wu et~al.(2023)Wu, He, Gao et~al.}]{GCEVT}
Wu, H.; He, Z.; Gao, M.; et~al. 2023.
\newblock GCEVT: Learning Global Context Embedding for Vehicle Tracking in Unmanned Aerial Vehicle Videos.
\newblock \emph{IEEE Geoscience and Remote Sensing Letters}, 20: 1--5.

\bibitem[{Wu et~al.(2025)Wu, Sun, Ji, and Kuang}]{wu2024temporal}
Wu, H.; Sun, H.; Ji, K.; and Kuang, G. 2025.
\newblock Temporal-Spatial Feature Interaction Network for Multi-Drone Multi-Object Tracking.
\newblock \emph{IEEE Transactions on Circuits and Systems for Video Technology}, 35(2): 1165--1179.

\bibitem[{Yang et~al.(2024)Yang, Han, Yan, Zhang, Qi, Lu, and Wang}]{yang2024hybrid}
Yang, M.; Han, G.; Yan, B.; Zhang, W.; Qi, J.; Lu, H.; and Wang, D. 2024.
\newblock Hybrid-sort: Weak cues matter for online multi-object tracking.
\newblock In \emph{Proceedings of the AAAI conference on artificial intelligence}, 6504--6512.

\bibitem[{Yao et~al.(2025)Yao, Peng, He, Peng, Chen, Chi, Liu, and Benediktsson}]{yao2024mm-tracker}
Yao, M.; Peng, J.; He, Q.; Peng, B.; Chen, H.; Chi, M.; Liu, C.; and Benediktsson, J.~A. 2025.
\newblock MM-Tracker: Motion Mamba for UAV-platform Multiple Object Tracking.
\newblock In \emph{Proceedings of the AAAI Conference on Artificial Intelligence}, 9409--9417.

\bibitem[{Yao et~al.(2023)Yao, Wang, Peng, Chi, and Liu}]{yao2023folt}
Yao, M.; Wang, J.; Peng, J.; Chi, M.; and Liu, C. 2023.
\newblock Folt: Fast multiple object tracking from uav-captured videos based on optical flow.
\newblock In \emph{Proceedings of ACM International Conference on Multimedia}, 3375--3383.

\bibitem[{Yi et~al.(2024)Yi, Luo, Luo, Huang, Wu, Hu, and Hao}]{yi2024ucmctrack}
Yi, K.; Luo, K.; Luo, X.; Huang, J.; Wu, H.; Hu, R.; and Hao, W. 2024.
\newblock Ucmctrack: Multi-object tracking with uniform camera motion compensation.
\newblock In \emph{Proceedings of the AAAI conference on artificial intelligence}, 6702--6710.

\bibitem[{Zhang et~al.(2022)Zhang, Sun, Jiang, Yu, Weng, Yuan, Luo, Liu, and Wang}]{zhang2022bytetrack}
Zhang, Y.; Sun, P.; Jiang, Y.; Yu, D.; Weng, F.; Yuan, Z.; Luo, P.; Liu, W.; and Wang, X. 2022.
\newblock Bytetrack: Multi-object tracking by associating every detection box.
\newblock In \emph{European conference on computer vision}, 1--21.

\bibitem[{Zhang et~al.(2021)Zhang, Wang, Wang, Zeng, and Liu}]{zhang2021fairmot}
Zhang, Y.; Wang, C.; Wang, X.; Zeng, W.; and Liu, W. 2021.
\newblock Fairmot: On the fairness of detection and re-identification in multiple object tracking.
\newblock \emph{International journal of computer vision}, 129: 3069--3087.

\bibitem[{Zheng et~al.(2025)Zheng, He, Chen, Zhang, Qu, and Wang}]{DFA-MOT}
Zheng, Y.; He, C.; Chen, X.; Zhang, H.; Qu, T.; and Wang, D. 2025.
\newblock DFA-MOT: A Dynamic Field-Aware Multi-Object Tracking Framework for Unmanned Aerial Vehicles.
\newblock \emph{IEEE Transactions on Circuits and Systems for Video Technology}, Early Access: 1--13.

\bibitem[{Zhu et~al.(2025)Zhu, Wang, Li, Tang, Gu, and Huang}]{zhu2025vtmot}
Zhu, Y.; Wang, Q.; Li, C.; Tang, J.; Gu, C.; and Huang, Z. 2025.
\newblock Visible--thermal multiple object tracking: Large-scale video dataset and progressive fusion approach.
\newblock \emph{Pattern Recognition}, 161: 111330.

\bibitem[{Zhuang et~al.(2024)Zhuang, Wang, Chen, Liu, Luo, and Tan}]{zhuang2024robust}
Zhuang, Z.; Wang, Z.; Chen, S.; Liu, L.; Luo, H.; and Tan, M. 2024.
\newblock Robust 3d semantic occupancy prediction with calibration-free spatial transformation.
\newblock \emph{arXiv preprint arXiv:2411.12177}.

\end{thebibliography}

\newpage

\section{-Supplementary Material-}
In this work, we propose AMOT, an advanced multi-object tracker built upon the joint detection and embedding (JDE) architecture~\cite{zhang2021fairmot}, augmented with two novel components: the appearance-motion consistency (AMC) matrix and the motion-aware track continuation (MTC) module. 
AMOT is specifically designed to enhance the robustness and accuracy of identity assignment under challenging UAV-captured videos. 
Comprehensive experiments and qualitative evaluations on UAV benchmarks are included in the supplementary material.

\subsection{Details of Hyperparameter Settings}
We conduct detailed ablation studies to analyze the impact of the hyperparameter settings used in the AMC matrix and the MTC module.
\subsubsection{Scale Factor $\sigma$ in the AMC Matrix.} 
\begin{figure}[!t]
    \centering
    \includegraphics[width=1.0\linewidth]{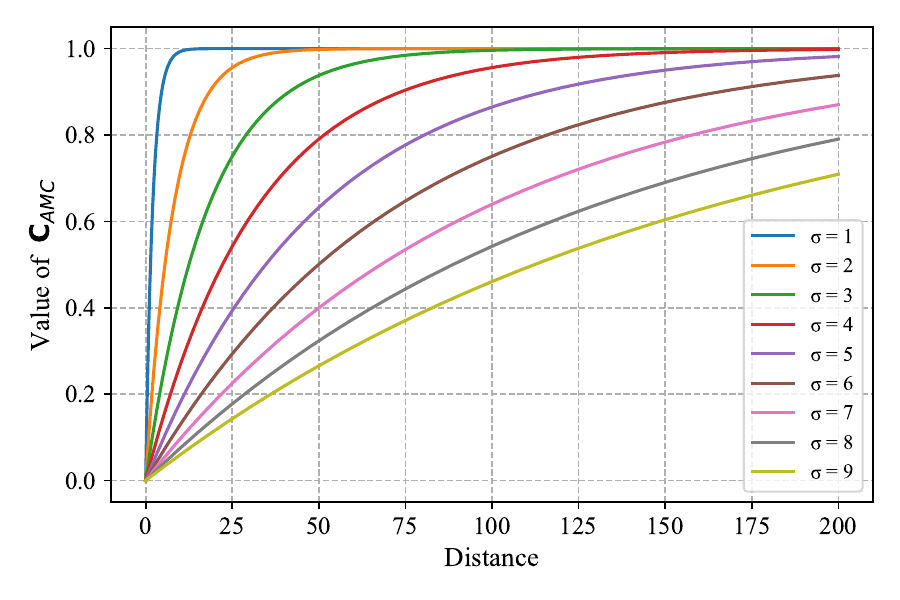}
    \caption{Impact of $\sigma$ on the sensitivity of $\mathbf{C}_{AMC}$ as the distance increases.}
    \label{fig:sigma}
\end{figure}

Here, we construct the AMC matrix $\mathbf{C}_{AMC}$ using a Gaussian kernel that integrates both forward and backward spatial distances, defined as:
\begin{align}
    \mathbf{C}_{AMC}(i,j) = 1 - \exp\left(-\frac{\mathbf{D}_{f}^{\top}(i,j) + \mathbf{D}_{b}(i,j)}{2\sigma^2}\right),
\end{align}
where $\mathbf{D}_{f}$ and $\mathbf{D}_{b}$ represent the forward and backward spatial distances, respectively, $\sigma$ is a scale factor that controls the spatial sensitivity. 
We employ a Gaussian kernel to normalize the bi-directional spatial distance into the range [0,1], thereby constructing a smooth and bounded affinity matrix. 
As illustrated in Figure~\ref{fig:sigma}, a smaller $\sigma$ (e.g., $\sigma=1$ or $\sigma=2$) results in a sharper decay of $\mathbf{C}_{AMC}$ with increasing spatial distance, reflecting higher sensitivity to spatial proximity. 
While a larger $\sigma$ (e.g., $\sigma=8$ or $\sigma=9$) leads to a smoother decay, indicating reduced sensitivity. 
Different values of $\sigma$ reflect varying sensitivities to spatial proximity. 

To further examine the impact of $\sigma$, we analyze the association performance under different settings, as reported in Table~\ref{sigma_result}. 
The IDF score exhibits an increasing trend with larger $\sigma$ values at first, but deteriorates beyond a certain point due to reduced sensitivity to spatial proximity. 
The optimal balance is achieved when $\sigma = 5$, which yields the highest IDF and a competitive number of identity switches (IDs). Thus, we adopt $\sigma = 5$ as the default setting in all experiments.

\begin{table}[t]
\small 
\setlength{\tabcolsep}{1.65mm} 
\centering
\begin{tabular}{c|ccccccc}
    \toprule
    $\sigma$  &IDF1$\uparrow$ &MOTA$\uparrow$ &MT$\uparrow$ &ML$\downarrow$ &FP$\downarrow$ &FN$\downarrow$ &IDs$\downarrow$\\
    \midrule
    1 &74.1 &55.1 &793 &142 &78727 &73949 &338\\
    2 &74.2 &55.1 &793 &142 &78821 &73930 &311\\
    3 &74.4 &55.1 &793 &142 &78925 &73900 &276\\
    4 &74.6 &55.1 &794 &142 &78931 &73907 &260\\
    % \rowcolor[HTML]{E8E8E8}
    5 &74.7 &55.1 &794 &142 &78928 &73968 &272\\
    6 &74.7 &55.1 &793 &142 &78723 &74151 &288\\
    7 &74.5 &55.1 &792 &143 &78346 &74256 &329\\
    8 &74.0 &55.1 &793 &144 &78161 &74353 &356\\
    9 &73.9 &55.2 &792 &144 &77619 &74622 &381\\
    10 &73.5 &55.3 &789 &146 &77052 &74883 &401\\
     \bottomrule
\end{tabular}
\caption{Impact of the scale factor $\sigma$ on the association performance.}  
\label{sigma_result}
\end{table}

\subsubsection{Predefined Threshold $\lambda$ in MTC Module.} 
\begin{figure}[!t]
    \centering
    \includegraphics[width=1.0\linewidth]{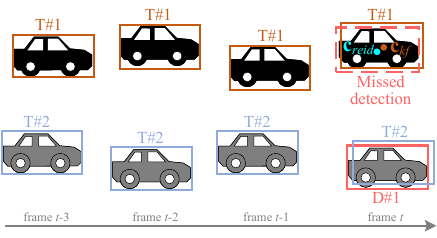}
    \caption{Overview of the MTC module. 
    At frame $t$, track T\#2 could be associated with detection D\#1, but track T\#1 remains unmatched due to a missed detection. 
    Subsequently, we determine whether to reactivate T\#1 by comparing its appearance-guided predicted center $\mathbf{c}_{reid}$ with the Kalman-based predicted center $\mathbf{c}_{kf}$.}
    \label{fig:model2}
\end{figure}

As shown in Figure~\ref{fig:model2}, we propose the MTC module to determine whether to reactivate unmatched tracks by computing the Euclidean distance $d$ between the appearance-guided predicted center $\mathbf{c}_{reid}$ with the Kalman-based predicted center $\mathbf{c}_{kf}$. This distance is formulated as:
\begin{align}
    d &= \left\| \mathbf{c}_{reid}-\mathbf{c}_{kf}\right\|_2.
\end{align} 
Then, we compare $d$ with a predefined threshold $\lambda$ to determine the activation of unmatched tracks. 
In Table~\ref{lamda_result}, we present the impact of different
predefined threshold $\lambda$. 
Although increasing $\lambda$ leads to a reduction in false negatives (FN), it also introduces more false positives (FP). 
Nevertheless, the best overall tracking performance is achieved at $\lambda = 3$, where both MOTA and IDF1 reach their highest values, indicating an effective balance among FP, FN, and IDs.
Consequently, we adopt $\lambda = 3$ as the default setting in all experiments. 

\begin{table}[t]
\small 
\setlength{\tabcolsep}{1.65mm} 
\centering
\begin{tabular}{c|ccccccc}
    \toprule
    $\lambda$  &IDF1$\uparrow$ &MOTA$\uparrow$ &MT$\uparrow$ &ML$\downarrow$ &FP$\downarrow$ &FN$\downarrow$ &IDs$\downarrow$\\
    \midrule
    1 &74.5 &55.1 &784 &143 &77651 &75132 &291\\
    2 &74.6 &55.1 &790 &142 &78469 &74402 &277\\
    % \rowcolor[HTML]{E8E8E8}
    3 &74.7 &55.1 &794 &142 &78928 &73968 &272\\
    4 &74.7 &55.0 &795 &142 &79266 &73807 &270\\
    6 &74.7 &54.9 &795 &142  &79660  &73676 &268\\
    8 &74.6 &54.9  &795 &142 &79830 &73645 &268\\
    10 &74.6 &54.9 &795 &142 &79911 &73616 &269\\
    15 &74.5 &54.8 &795 &142 &80126 &73574 &271\\
    20 &74.5 &54.8 &796 &142 &80268 &73539 &271\\
    40 &74.4 &54.5 &798 &142 &81440 &73388 &271\\
     \bottomrule
\end{tabular}
\caption{Impact of the predefined threshold $\lambda$ on the tracking performance.}
\label{lamda_result}
\end{table}

\begin{table}[t]
\small 
\setlength{\tabcolsep}{1.3mm} 
\centering
\begin{tabular}{c|ccc|ccc}
    \toprule
    \multirow{2}{*}{\centering Cost matrix} &\multicolumn{3}{c|}{VisDrone2019} &\multicolumn{3}{c}{UAVDT}\\
    \cline{2-7}
    &IDF1$\uparrow$  &MOTA$\uparrow$ &IDs$\downarrow$ &IDF1$\uparrow$  &MOTA$\uparrow$ &IDs$\downarrow$  \\
    \midrule
    % \rowcolor[HTML]{E8E8E8}
   $\mathbf{C}_{uni}^{\scriptscriptstyle(1)}$  &61.4 &46.0 &1063 &74.7 &55.1 &272\\  
   $\mathbf{C}_{uni}^{\scriptscriptstyle(2)}$  &56.8 &45.3 &2480 &73.8 &55.1 &446\\
   $\mathbf{C}_{uni}^{\scriptscriptstyle(3)}$  &61.2 &46.1 &1138 &74.3 &55.0 &305 \\
    \bottomrule
\end{tabular}
\caption{Impact of different combinations of the unified cost matrix.}
\label{cost_ablation_result}
\end{table}

\begin{figure}[!t]
    \centering
    \includegraphics[width=1.0\linewidth]{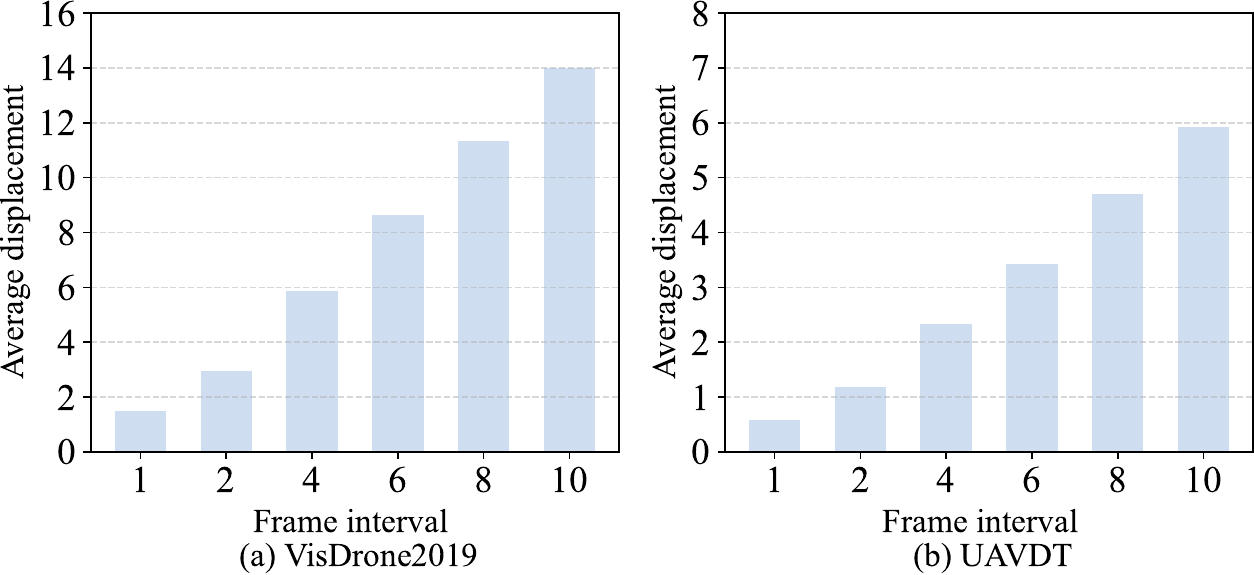}
    \caption{
    The average object displacement is calculated over all sequences and objects under different frame intervals.}
    \label{fig:dis}
\end{figure}

\subsubsection{Unified Cost Matrix Formulations.} 
Here, we explore several formulations for combining different cost matrices, including the AMC matrix $\mathbf{C}_{AMC}$, IoU matrix $\mathbf{C}_{IOU}$, and appearance similarity matrix $\mathbf{C}_{App}$, and ultimately construct a unified cost matrix $\mathbf{C}_{uni}$ for data association. 
$\mathbf{C}_{uni}$ aims to leverage complementary information from these individual matrices to better measure the affinity between detections and tracks. 
Besides $\mathbf{C}_{uni}^{\scriptscriptstyle(1)}$, we also investigate the following two formulations:
\begin{align}
    \mathbf{C}_{uni}^{(2)} = 1-(1-\mathbf{C}_{IOU}) \cdot (1- \mathbf{C}_{AMC} \cdot \mathbf{C}_{App}),
\end{align} 

\begin{align}
    \mathbf{C}_{uni}^{(3)} = 1-(1-\mathbf{C}_{AMC} \cdot \mathbf{C}_{IOU}) \cdot (1- \mathbf{C}_{AMC} \cdot \mathbf{C}_{App}).
\end{align} 
As shown in Table~\ref{cost_ablation_result}, $\mathbf{C}_{uni}^{\scriptscriptstyle(1)}$ consistently achieves better tracking performance, demonstrating the effectiveness of its design.
While $\mathbf{C}_{uni}^{\scriptscriptstyle(2)}$ demonstrates worse performance, this may be due to the AMC matrix failing to compensate for the instability of the IoU matrix. 
Such results indicate that the AMC matrix effectively models the joint relationship between visual similarity and spatial coherence.

\subsection{Further Evaluation on Large Displacements}
In Figure~\ref{fig:dis}, we present the average object displacement calculated over all sequences and objects in the VisDrone2019 and UAVDT test sets under increasing frame intervals.  
As the frame interval increases, the average displacement exhibits a clear upward trend, indicating that longer temporal gaps introduce greater motion uncertainty and pose challenges for multi-object trackers.
The subsection titled \textbf{\textit{``Robustness to Large Displacements''}} in the manuscript has validated the robustness of our AMOT in addressing such challenging scenarios.
\end{document}